\documentclass[table, dvipsnames]{article} %
\usepackage{collas2024_conference,times}
\usepackage{easyReview}
\usepackage{amssymb}
\usepackage{bm}
\usepackage{multirow}
\usepackage{multicol}
\usepackage{arydshln}
\usepackage{graphicx}
\usepackage{subcaption}
\usepackage{wrapfig}
\usepackage{xspace}
\usepackage{algorithm}
\usepackage{algpseudocode}
\usepackage{soul}
\usepackage{subcaption}
\usepackage{float}
\usepackage{tikz}
\usepackage{calc}
\usepackage{booktabs}
\usepackage{makecell}

\usepackage{amsmath,amsfonts,bm}

\def\eqref#1{equation~\ref{#1}}

\def\1{\bm{1}}

\def\vg{{\bm{g}}}
\def\vh{{\bm{h}}}

\def\vk{{\bm{k}}}

\def\vp{{\bm{p}}}

\def\vs{{\bm{s}}}

\def\vx{{\bm{x}}}
\def\vy{{\bm{y}}}

\DeclareMathAlphabet{\mathsfit}{\encodingdefault}{\sfdefault}{m}{sl}
\SetMathAlphabet{\mathsfit}{bold}{\encodingdefault}{\sfdefault}{bx}{n}

\usepackage{hyperref}
\hypersetup{
    colorlinks=true,
    linkcolor=red,
    filecolor=magenta,
    urlcolor=blue,
    citecolor=purple,
    pdftitle={Overleaf Example},
    pdfpagemode=FullScreen,
    }

\title{Less is more: Summarizing Patch Tokens for efficient Multi-Label Class-Incremental Learning}

\author{Thomas De Min\thanks{Corresponding Author} \\
University of Trento \\
\texttt{thomas.demin@unitn.it}
\And
Massimiliano Mancini \\
University of Trento
\And
Stéphane Lathuilière \\
LTCI, Télécom Paris, \\
Institut Polytechnique de Paris
\AND
Subhankar Roy \\
University of Aberdeen
\And
Elisa Ricci \\
University of Trento \\
Fondazione Bruno Kessler
}

\def\checkmark{\tikz\fill[scale=0.4](0,.35) -- (.25,0) -- (1,.7) -- (.25,.15) -- cycle;} 
\def\scalecheck{\resizebox{\widthof{\checkmark}*\ratio{\widthof{x}}{\widthof{\normalsize x}}}{!}{\checkmark}}

\newcommand{\name}{\textsc{multi-lane}\xspace}

\definecolor{colorselectors}{RGB}{250, 217, 213}
\definecolor{colorsummarized}{RGB}{255, 230, 204}
\definecolor{colorcls}{RGB}{177, 221, 240}

\definecolor{colormethod}{RGB}{200, 231, 250}
\newcommand{\colormethod}{colormethod}
\newcommand{\rowmethod}{\rowcolor{\colormethod}}
\newcommand{\colorcil}{Apricot!40}
\newcommand{\rowcil}{\rowcolor{\colorcil}}
\newcommand{\colormlcil}{Purple!20}
\newcommand{\rowmlcil}{\rowcolor{\colormlcil}}
\newcommand{\inlineColorbox}[2]{\begingroup\setlength{\fboxsep}{1pt}\colorbox{#1}{\hspace{2pt}\vphantom{Ay}#2\hspace{2pt}}\endgroup}

\definecolor{my_green}{RGB}{16,119,51}

\definecolor{my_red}{RGB}{255,25,25}

\collasfinalcopy %

\begin{document}

\maketitle

\begin{abstract}

Prompt tuning has emerged as an effective rehearsal-free technique for class-incremental learning (CIL) that learns a tiny set of task-specific parameters (or \textit{prompts}) to instruct a pre-trained transformer to learn on a sequence of tasks. Albeit effective, prompt tuning methods do not lend well in the multi-label class incremental learning (MLCIL) scenario (where an image contains multiple foreground classes) due to the ambiguity in selecting the correct prompt(s) corresponding to different foreground objects belonging to multiple tasks. To circumvent this issue we propose to eliminate the prompt selection mechanism by maintaining task-specific \textit{pathways}, which allow us to learn representations that do not interact with the ones from the other tasks. Since independent pathways in truly incremental scenarios will result in an explosion of computation due to the quadratically complex multi-head self-attention (MSA) operation in prompt tuning, we propose to reduce the original patch token embeddings into \textit{summarized} tokens. Prompt tuning is then applied to these fewer summarized tokens to compute the final representation. Our proposed method \underline{Multi} \underline{L}abel class incremental learning via summarising p\underline{A}tch toke\underline{N} \underline{E}mbeddings (\name) enables learning disentangled task-specific representations in MLCIL while ensuring fast inference. We conduct experiments in common benchmarks and demonstrate that our \name achieves a new state-of-the-art in MLCIL.
Additionally, we show that \name is also competitive in the CIL setting. Source code available at \url{https://github.com/tdemin16/multi-lane}

\end{abstract}

\section{Introduction}
\label{sec:introduction}

Class-incremental learning (CIL)~\citep{masana2022class} aims to learn a classification model on a sequence of tasks (i.e. datasets with annotations for new classes) without forgetting previous knowledge and running into \textit{catastrophic forgetting} \citep{mccloskey1989catastrophic}. %
Standard CIL methods assume that an image contains an object from a single class. Whereas in practice, images rarely depict a single subject (\textit{e.g.}, an urban scene contains cars, traffic lights, people, etc.).  
To relax this assumption, multi-label class incremental learning (MLCIL)~\citep{kim2020imbalanced, dong2023knowledge} aims to correctly classify an image into multiple classes, that may be introduced across different tasks.

Different from standard CIL, MLCIL is characterized by two additional problems.
\textit{First}, when updating the model, annotations are available only for the objects learned at the current task.
As a result, if {such} images contain previously learned categories, they become negative samples for the old classes, increasing forgetting~\citep{dong2023knowledge}.
\textit{Second}, multi-label classification datasets are governed by long-tailed distributions~\citep{kim2020imbalanced}.
To prevent the forgetting of old knowledge, rehearsal-based methods store and replay samples of old training sessions~\citep{kim2020imbalanced, dong2023knowledge}. While this ensures positive training instances for both old and new classes, storing examples might not be possible due to privacy regulations~\citep{zhu2021prototype,wang2022learning}.
\begin{figure}[h]
    \centering
    \includegraphics[width=\linewidth]{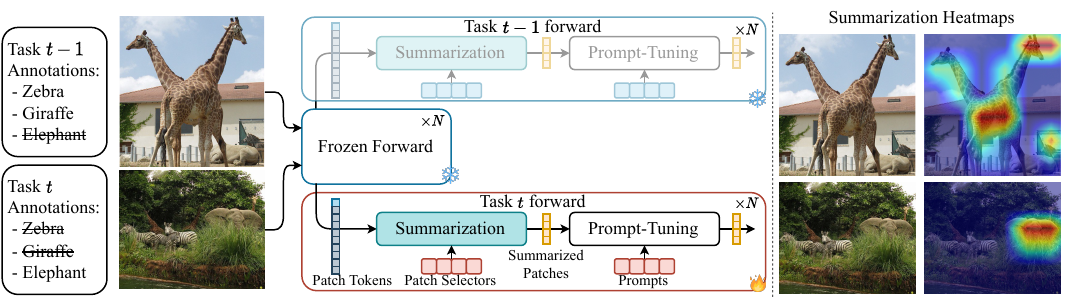}
    \caption{\textbf{A schematic overview of MLCIL and \name}. Annotations in MLCIL are incomplete, as shown with a strikethrough, even though images can depict objects from such classes. In \name, we learn task-specific subnetworks. Inside each subnetwork, Patch Selectors attend to the patch tokens and summarize them into fewer summarized patches. These summarized patches are used for prompt tuning, thus, reducing the computational complexity in each task forward. On the right, we visualize the attention heatmaps produced by Patch Selectors.}
    \label{fig:teaser}
    \vspace{-5mm}
\end{figure}

In CIL, parameter-efficient fine-tuning (PEFT) techniques~\citep{wang2022dualprompt,wang2022learning, smith2023coda, de2023effectiveness} are an effective alternative to rehearsal-based methods \citep{rebuffi2017icarl,buzzega2020dark}. In particular, prompt tuning \citep{jia2022visual,wang2022learning} learns a tiny set of task-specific parameters (or \textit{prompts}) attached to a frozen pre-trained Vision Transformer (ViT) \citep{50650}. 
These prompts are stored and selected to condition the final representation from the pre-trained backbone, adapting it to the task at hand. Although more effective than the rehearsal-based CIL methods, the success of prompt tuning heavily relies on selecting the right prompt(s) at inference time \citep{wang2022dualprompt}. The prompt selection issue is exacerbated in the MLCIL setting as multiple prompts need to be selected corresponding to foreground objects learned across tasks (e.g. in Fig.~\ref{fig:teaser}, the bottom image contains \textit{zebra}, learned in task $t-1$, and \textit{elephant}, learned in task $t$).
Thus, if prompts are not selected carefully, the model may misinterpret foreground and background information.
This makes prompt tuning a non-trivial approach for MLCIL (further details in Appx.~\ref{sec:misinterpret}).
As prompt tuning does not lend well to the MLCIL setting, existing methods directly address these two issues by allocating ad-hoc memory buffers \citep{kim2020imbalanced, liang2022optimizing} and by pseudo-labeling new images for old classes \citep{dong2023knowledge}.

In this work, we argue that prompt tuning can indeed be an effective solution for MLCIL if prompts across tasks do not interact with each other. To this end, we present our method, \underline{Multi} \underline{L}abel class incremental learning via summarising p\underline{A}tch toke\underline{N} \underline{E}mbeddings (\name) for MLCIL.
In detail, we propose to eliminate cross-task prompt interaction and prompt selection at inference time by learning task-specific \textit{pathways} or subnetworks (see Fig.~\ref{fig:teaser}). 
Such a design allows a more equal representation capacity for all classes across tasks while avoiding future samples becoming negatives for old classes. 
Albeit promising, a na\"ive implementation of such \textit{pathways} would linearly scale the expensive multi-head self-attention (MSA) operations as the number of tasks increases, making inference in truly lifelong scenarios infeasible. 
As a remedy, we propose to summarize the patch token embeddings into fewer \textit{summarized} patch embeddings. 
We do so by learning a set of tokens called \textit{Patch Selectors} (see Sec.~\ref{sub:compression} for details) that attend to relevant regions of the image for the current task and summarize the semantic content (see Fig.~\ref{fig:teaser} right for regions relevant to the tasks). 
These summarized tokens are then fed to the prompt tuning layers for learning task-specific representation. 
As the summarized tokens are fewer in number ($\sim 1-20$ in our experiments) than the original patch tokens (196 patch tokens with a 14 $\times$ 14 patchification in ViT-B/16), the number of MSA operations is drastically reduced due to quadratic nature of the MSA complexity.
This allows us to sustain as many task-specific subnetworks as the number of tasks while ensuring a competitive inference speed. 
Moreover, \name is trained with only a classification loss and does not require any common regularization losses \citep{li2017learning} to prevent forgetting.
During inference, the test image is forwarded through all the subnetworks in the form of summarized patches, and the resulting task logits are concatenated to yield the final class prediction.

Experiments on MS-COCO~\citep{lin2014microsoft} and VOC2007~\citep{pascal-voc-2007} datasets demonstrate that \name achieves a new state-of-the-art in the MLCIL setting while keeping the computational requirements similar to standard ViT. Unlike existing MLCIL methods \citep{dong2023knowledge} that still rely on the rehearsal for good performance, \name does not require a memory buffer.
Additional experiments on CIFAR-100~\citep{krizhevsky2009learning} and ImageNet-R~\citep{hendrycks2021many} show that our approach is also competitive in the standard CIL scenario.

In summary, our \textbf{contributions} are the following: (\textbf{i}) We propose \name that, to the best of our knowledge, is the first prompt-based framework for the challenging MLCIL setting; (\textbf{ii}) We introduce Patch Selectors, a simple yet effective patch summarization technique, that actualizes task-specific subnetworks while preventing the computational complexity from growing; and (\textbf{iii}) \name achieves a new state-of-the-art in MLCIL while not requiring rehearsal. Contrary to previous MLCIL approaches, \name is also competitive in the CIL setting.

\section{Related Works}
\label{sec:related}

\paragraph{Continual Learning.}
Continual Learning methods can be broadly categorized into \textit{regularization-based}, \textit{rehearsal-based}, and \textit{parameter-isolation} approaches~\citep{de2021continual}. 
Regularization-based methodologies tackle the problem of catastrophic forgetting via a regularization term that constrains the weights or activations of the model from deviating from the old configuration~\citep{li2017learning, kirkpatrick2017overcoming}.
Although regularization-based approaches improve over the sequential fine-tuning baseline, their effectiveness is limited. 
By allowing small portions of the past samples to be stored in a memory and replayed during learning on the current task, rehearsal-based methods~\citep{rebuffi2017icarl, buzzega2020dark,prabhu2020gdumb} show impressive results, comparable to joint training.
However, relying on a replay buffer raises privacy-related concerns about stored images~\citep{wang2023comprehensive} and accrues computation as the task sequence grows.
Finally, parameter isolation methods maintain task-specific subnetworks inside the model, where parameters across tasks do not interact with each other \citep{mallya2018packnet,mallya2018piggyback}. Very recently, ever since the advent of strong pre-trained ViT models, there has been a redux of parameter isolation methods that are based on visual prompt tuning \citep{jia2022visual} and have demonstrated performance on-par or even better than rehearsal-based methods. In detail, a tiny set of learnable parameters or prompts are trained for each task and used to condition the final representation from the ViT \citep{wang2022dualprompt,wang2022learning,smith2023coda}. \name falls into the family of prompt tuning-based CIL methods but deviates from the existing related works because it eliminates prompt selection during inference and drastically reduces computation in MSA blocks via patch summarization.

\paragraph{Multi-Label Class-Incremental Learning.}
While multi-label classification is a fairly mature research field \citep{wang2017multi,ridnik2021asymmetric,liu2021query2label,lanchantin2021general}, it has not received adequate attention in the incremental setting MLCIL \citep{dong2023knowledge}. In detail, KRT \citep{dong2023knowledge} addresses the MLCIL problem by dynamically computing pseudo-labels for the old classes and by learning a cross-attention module to transfer knowledge from the old classes onto the new ones. Related to MLCIL, Partitioning Reservoir Sampling (PRS)~\citep{kim2020imbalanced} tackles online incremental learning in the context of multi-label classification by designing a reservoir sampling strategy for the long-tailed distribution problem. Similarly, OCDM~\citep{liang2022optimizing} optimizes the class distribution in memory to improve the performance over PRS. Different from the existing MLCIL approaches, \name is based on prompt tuning and does not require storing samples from past tasks, thus, it is the first rehearsal-free MLCIL method.

\paragraph{Token reduction in ViT.}
As the MSA in ViT scales quadratically in the number of patch tokens, token sparsification approaches have focused on reducing the input token sequence length by either token pruning or token merging. Pruning-based methods aim to reduce the token sequence length by removing either a predetermined number of tokens \citep{kong2021spvit,liang2022not} or a dynamically adaptive number of tokens \citep{pan2021ia,yin2022adavit}.
Whereas, merging-based methods focus on reducing the sequence length by combining tokens \citep{ryoo2021tokenlearner,bolya2022tome,marin2023token}. In particular, TokenLearner~\citep{ryoo2021tokenlearner} learns an MLP that outputs a spatial weight map over image features, which is then used to compress information by averaging the weighted spatial features. The patch summarization in \name is similar in spirit to TokenLearner, except we do not train an MLP, but task-specific learnable tokens (or Patch Selectors) learn to summarize the intermediate patch tokens into fewer summarized tokens via dot product attention. To the best of our knowledge, we are the first to incorporate the idea of token reduction in prompt tuning for continual learning, thereby bridging the two communities.

\section{Method}
\label{sec:method}
In this Section, we first formalize the multi-label class-incremental learning problem (Sec.~\ref{sub:problem}) and provide background on continual learning via prompt-tuning and its limitations (Sec.~\ref{sub:clpt}).
Finally, we delve into the motivations behind our method \name and describe its components (Sec.~\ref{sub:compression}).

\subsection{Problem formulation}
\label{sub:problem}
The goal of multi-label class incremental learning (MLCIL) is to classify multiple objects present in an input image, given that the model is incrementally trained on datasets annotated for only a subset of classes.

Formally, we aim to learn a function $f_\theta:\mathcal{X}\rightarrow \mathcal{Y}$, parametrized by $\theta$, mapping images in the space $\mathcal{X}$ to binary vectors in the space $\mathcal{Y}$, where $\mathcal{Y}=\{0,1\}^k$ and $k$ is the number of classes. %
In the incremental learning scenario, the training dataset is defined as $\mathcal{D} = \{\mathbf{D}^1, \dots, \mathbf{D}^T\}$, where $T$ is the number of training steps and $\mathbf{D}^t$ is the dataset available at step $t$.
Each dataset represents a collection of image-label pairs $\mathbf{D}^t = \{(\vx^t_i, y^t_i)\}_{i=1}^{n_t}$, where $\vx_i^t$ and $y_i^t$ are respectively the $i$-th image and labels for task $t$. %
Note that, in each training step $t$, we have annotations for a different set of semantic categories $\mathcal{Y}^t$, with no overlap across training steps, i.e. $\mathcal{Y}^t \cap \mathcal{Y}^u = \varnothing$ for each %
$t \neq u$.  As a consequence, a training image for a task $t$ has ground-truth positive/negative annotations only for the subset $\mathcal{Y}^t$ (further details in Appx.~\ref{sec:transition}). The final output space is thus defined as $\mathcal{Y} = \bigcup_{t=1}^T \mathcal{Y}^t$. %

\subsection{Preliminaries: Continual Learning via Prompt-Tuning}
\label{sub:clpt}
We follow a family of CIL methods that utilize prompt tuning to circumvent the need to store past samples \citep{wang2022dualprompt,wang2022learning,smith2023coda}. The main idea in prompt tuning is to prepend extra learnable tokens (or \textit{prompts}) to the input sequences at intermediate layers of the ViT, that are optimized to learn new tasks while keeping the weights of the ViT backbone frozen. These prompts then serve as \textit{instructions}, encoding the task-related information, to modulate the pre-trained representations for downstream tasks.

In the context of ViT \citep{50650}, given an input image $\vx$, let $\vh \in \mathbb{R}^{L \times D}$ be the sequence-like output embedding from the input Embedding block, where $L$ is the length of the sequence and $D$ is the embedding dimension. To learn the current task $t$, the ViT backbone is kept frozen, and the task-specific learnable prompt parameters $\vp_t \in \mathbb{R}^{L_{\vp} \times D}$, having sequence length $L_{\vp}$ and dimension $D$, are prepended to $\vh$ to form the extended embedding features. Formally, the input to the $i$-th multi-head self-attention (MSA) layer \citep{vaswani2017attention} $\vh^{(i)}$ is modulated by a \textit{prompting function} as:

\vspace{-6mm}
\begin{equation}
\label{eqn:prompt-tuning}
    \vh^{(i)}_{\text{mod}} = f_{\text{prompt}}(\vp_t, \vh^{(i)}),
\end{equation}
where $f_{\text{prompt}}$ is an MSA transformation function determining how the prompts are attached to the hidden embeddings. Specifically, applying $f_{\text{prompt}}$ can be viewed as modifying the generic inputs $\vh \in \mathbb{R}^{L \times D}$ to the MSA layer. The input $\vh$ is projected into query, key, and values, denoted as $\vh_Q, \vh_K$ and $\vh_V$, by projection matrices $W^Q$, $W^K$, and $W^V$. The MSA layer is defined as:

\vspace{-5mm}
\begin{align}
    & \text{MSA}(\vh_Q, \vh_K, \vh_V) = \text{Concat}(\text{h}_1, \dots, \text{h}_\text{m}) W^{O} \nonumber\\
    & \text{where} \: \text{h}_\text{i} = \text{Attention}(\vh_Q, \vh_K, \vh_V), \nonumber
\end{align}
where $W^O$ is a projection matrix and $\text{m}$ is the number of attention heads. Using this formulation of MSA, the $f_{\text{prompt}}$ that implements prefix-tuning \citep{wang2022dualprompt} in particular is defined as:

\vspace{-5mm}
\begin{align}
    f_\text{prompt}(\vp_t, \vh^{(i)}) = \text{MSA}(\vh^{(i)}_{Q}, [\vp_{t, K};\vh^{(i)}_K], [\vp_{t, V};\vh^{(i)}_V]),
\end{align}
where $\vp_t$ is split into $\vp_{t,K}, \vp_{t,V} \in \mathbb{R}^{L_{\vp}/2 \times D}$, and then prepended to $\vh^{(i)}_{K}$ and $\vh^{(i)}_{V}$, respectively, while keeping $\vh^{(i)}_{Q}$ unchanged \citep{wang2022dualprompt}. Additionally, each task-specific prompt $\vp_t$ is associated with a learnable key $\vk_t \in \mathbb{R}^D$. The keys are trained to match the feature of the input instance, obtained with a query function $q(\cdot)$, using a matching loss. During inference, $q(\cdot)$ first selects the appropriate key $\vk_t$ associated with the prompt $\vp_t$. The corresponding $\vp_t$ is then used to obtain the final modulated representation from the ViT backbone.

The prompt tuning methods, albeit effective for CIL, are not well-suited for MLCIL because images with similar semantics may belong to different tasks (foreground may become background in future tasks) \citep{dong2023knowledge}, thus, query-key matching may fail during prompt selection. If selected prompts are wrong, they may misinterpret foreground information for irrelevant features.

\subsection{Multi-Label Class Incremental Learning via Summarising Patch Token Embeddings}
\label{sub:compression}
\begin{figure}[t]
    \centering
    \includegraphics[width=0.90\linewidth]{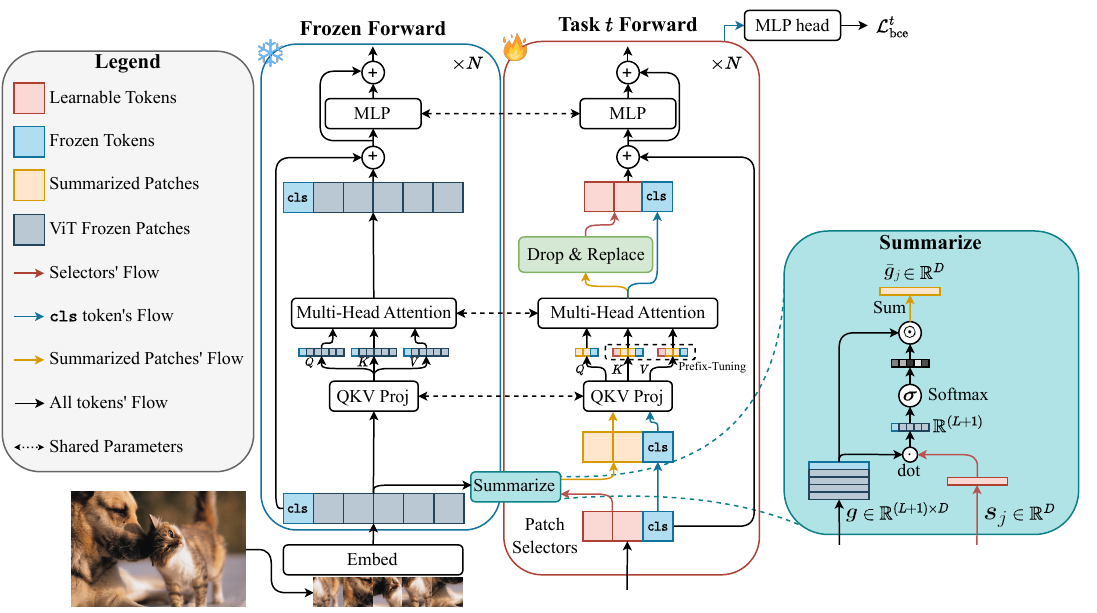}
    \caption{\textbf{Training of \name at task $t$}. \inlineColorbox{colorselectors}{Patch Selectors} summarize patch token embeddings into fewer specialized tokens, reducing the computational complexity of the multi-head attention strongly. The MSA is computed between \inlineColorbox{colorsummarized}{summarized tokens} and the \inlineColorbox{colorcls}{\texttt{[CLS]} token}, and it is fine-tuned with prompt-tuning. After the computation of the MSA, summarized patches are dropped and only Patch Selectors and \texttt{[CLS]} token are propagated.}
    \label{fig:method}
    \vspace{-5mm}
\end{figure}

In order to avoid the critical drawbacks of prompt tuning methods, we design \name that entirely eliminates the need to select prompts with a query function. Instead, we create task-dependant \textit{pathways} (or subnetworks), where each subnetwork implements prompt tuning with task-specific data. While this solution can eliminate interaction between tasks, and handle long-tailed distributions, it introduces computation costs during inference that scales linearly with the number of tasks, making it unfit for truly lifelong scenarios. To recall, the bulk of the computation in ViT is incurred in the MSA layers, causing the compuational complexity to grow quadratically in the token length $O(L^2)$ \citep{vaswani2017attention}. Given we intend to have a separate forward pass for each task, the complexity rises to the order $O(TL^2)$, which scales linearly with $T$. 

To still benefit from parallel subnetworks and yet keep computation under check, we propose the idea of \textit{Patch Summarization}. In a nutshell, we introduce trainable tokens called \textit{Patch Selectors} that learn to attend to the regions of the image, relevant to the current task, and \textit{summarize} the $L$ patch tokens into fewer summarized patch tokens. Unlike L2P \citep{wang2022learning} or DualPrompt \citep{wang2022dualprompt}, the prompt tuning in \name operates on top of the summarized patch tokens, rather than the full sequence of patch tokens. Next, we describe Patch Selectors in detail, followed by how prompt tuning is realized.

\paragraph{Patch Selectors.} Let us denote the learnable Patch Selectors for a task $t$ as $\vs_t \in \mathbb{R}^{L_\vs \times D}$, having sequence length $L_\vs$ and dimension $D$. We ensure that the sequence length of the Patch Selectors $L_\vs$ is much smaller than the patch embedding length $L$. We also keep a copy of the class token ($\texttt{[CLS]}$), which is forwarded along with the Patch Selectors. We initialize one $\texttt{[CLS]}$ per task with pre-trained weights \citep{wang2022dualprompt,wang2022learning} as it allows for independent representations for each encountered task. At training time to compute the summarizations for each task, we maintain two forward passes (see Fig. \ref{fig:method}): (i) a \textit{frozen forward} where we obtain the intermediate representation of an image using a frozen ViT backbone, consisting of patch embeddings and the $\texttt{[CLS]}$ token embedding; and (ii) \textit{task forward} where the task-specific Patch Selectors $\vs_t$ compute the importance of the input patch and the $\texttt{[CLS]}$ token embeddings for the given task $t$. 
Let $\vg^{(l)} \in \mathbb{R}^{(L+1) \times D}$ be the intermediate representation of the $\texttt{[CLS]}$ and patch embeddings at the $l$-th transformer block. Dropping the layer index $l$ for the convenience of notation, the $j$-th Patch Selector for task $t$ summarizes the input embeddings $\vg$ into $\bar{\vg}$ as:

\vspace{-5mm}
\begin{align}
    \label{eqn:summarization}
    \bar{\vg}_{j} = \displaystyle \sum^{L+1}_{k=1} \bm{\alpha}_k \vg_k, \quad
    \text{where} \; \bm{\alpha} = \text{softmax} \left (\frac{\vs_j \vg^{\textrm{T}}}{\sqrt{D}} \right).
\end{align}
$\bm{\alpha}$ denotes the contributions of the embeddings $\vg$ corresponding to the Patch Selector $\vs_j$. Similarly, by concatenating the outputs of $L_\vs$ Patch Selectors, we obtain the summarized patches $\bar{\vg} \in \mathbb{R}^{L_\vs \times D}$. 

As, $L_\vs < L$, the computation in the MSA block of task-specific forward reduces to the order $O(L^2_\vs) \ll O(L^2)$. This reduction in computation allows us to afford $T$ task-specific forward passes, making the representation learned by the current task-specific parameters (or \textit{pathways}) independent of the previous tasks. These summarized patches are then fed to the MSA, which is discussed next. 

\paragraph{Fine-tuning.} After the patches have been summarized, we employ prefix-tuning \citep{wang2022dualprompt} to encode task-specific knowledge in the model. Following the notation used in Sec. \ref{sub:clpt}, the modulated representation presented to the $i$-th MSA layer is given as:

\vspace{-3mm}
\begin{equation}
    \vh^{(i)}_{\text{mod}} = f_\text{prompt}(\vp_t, \bar{\vg}^{(i)}).
\end{equation}
Compared to Eq. (\ref{eqn:prompt-tuning}), we use $\bar{\vg}^{(i)}$ as an input to the prompting function. Note that different from the previous prompt tuning-based methods \citep{wang2022dualprompt,wang2022learning}, in our formulation the sequence length of the input embeddings is greatly reduced, which reduces the computational overhead incurred in the MSA layers. As the internal representation changes during the forward pass, we compute the summarization before each MSA, capturing salient task-specific features at each step of the transformer.
Since tokens do not interact until the next MSA, we drop the summarized patches (Drop \& Replace in Fig.~\ref{fig:method}), propagating only the $\texttt{[CLS]}$ token and the patch Patch Selectors.
Algo.~\ref{alg:summarization} (Appx.~\ref{sec:pseudo}) shows the pseudocode of patch selection and fine-tuning.

The $\texttt{[CLS]}$ token conditioned by the task-specific parameters, emerging out from the final ViT block in the task forward, is used for learning a classifier. Formally, let $\hat{\vy}$ be the predicted sigmoid-normalized logits for classes in task $t$. The learning objective is Binary Cross-Entropy loss over the current task classes:

\vspace{-3mm}
\begin{equation}
    \label{eqn:bce}
    \mathcal{L}^t_{\text{bce}} = -\frac{1}{|\mathcal{Y}^t|}\sum_{i \in \mathcal{Y}^t} y_i\log \hat{y}_i + (1-y_i) \log(1-\hat{y}_i)
\end{equation}
where $i$ is the class index for classes in $\mathcal{Y}^t$, $y_i$ is the ground truth binary label and $\hat{y}_i$ is the corresponding model prediction.

\paragraph{Inference.}
\begin{wrapfigure}[14]{r}{0.25\textwidth}
    \vspace{-40pt}
    \includegraphics[width=\linewidth]{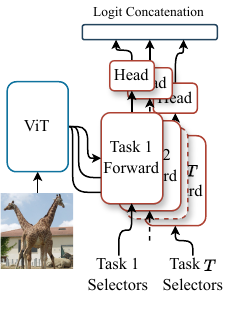}
    \caption{\name{} at inference time: all task forwards are run in parallel.}
    \label{fig:inference}
\end{wrapfigure}
During inference, the task identities of the image are not needed as \name is task agnostic.
In detail, the test image is forwarded through the frozen forward, while Patch Selectors are forwarded through $T$ task forward pathways (See Fig.~\ref{fig:inference}). 
The $\texttt{[CLS]}$ in each of the $T$ forwards ends in a specific classification head, which results in $T$ logits. 
The task-specific logits are then concatenated to yield the final prediction as $\hat{\bm{Y}} = [\hat{\vy}_1; \dots; \hat{\vy}_T]$, where $[\cdot;\cdot]$ denotes the concatenation operation. The final prediction is done by independent thresholding of each sigmoid-normalized logit, which is the standard procedure in multi-label and binary classification.

\section{Experiments}
\label{sec:experiments}
In this section, we present our experimental procedure (Sec.~\ref{sub:experimental}), and we compare \name with the current MLCIL state-of-the-art (Sec.~\ref{sub:comparison}).
We further evaluate our method by visualizing the summarization heatmaps generated by our Patch Selectors during summarization (Sec.~\ref{sub:qualitative}).
Following, we provide a complete ablation of the different components of our approach by removing the multiple \textit{pathways}, Drop \& Replace, and the pre-head normalization.
Additionally, we show how the performances of \name vary when changing the number of Patch Selectors or substituting them with Token Merging (Sec.~\ref{sub:ablations}).
To demonstrate that \name multiple \textit{pathways} are computationally sustainable, we compare the number or multiply-accumulate operations with the SOTA prompt-based approach.
Finally, we test our approach in two common benchmarks in CIL, showing that \name is also competitive in this setting (Sec.~\ref{sub:additional}).

\subsection{Experimental setting}
\label{sub:experimental}

\paragraph{Datasets.}
We follow previous works in the field~\citep{kim2020imbalanced, liang2022optimizing, dong2023knowledge} and test \name on MS-COCO~\citep{lin2014microsoft}, and VOC2007~\citep{pascal-voc-2007} datasets.
MS-COCO consists of 300k images ranging in 80 categories from the real-world domain.
Similarly, VOC2007 is a 10k image dataset annotated for 20 classes.
Both are initially meant for segmentation tasks, however, %
by following \citet{dong2023knowledge}, we adapted them to the multi-label class-incremental setting. 
We trained our model in two configurations for both datasets: a) MS-COCO B0-C10, the model was trained incrementally, and each task is annotated with 10 novel classes; b) MS-COCO B40-C10, we first trained on the first task composed of 40 classes, and then trained incrementally on the remaining 4; c) VOC2007 B0-C4, we trained incrementally on 5 tasks, each with 4 new classes; d) VOC2007 B10-C2, as for MS-COCO, we first trained on the first task composed of 10 classes, and then fine-tuned incrementally on the remaining 10, spanning 5 tasks.

\paragraph{Metrics.}
Following \citet{kim2020imbalanced} and \citet{dong2023knowledge}, %
we report the Average mean Average Precision across different tasks (Avg. mAP), and the mean Average Precision after training on the last task and evaluating on all seen classes (mAP).
We also report the class-wise F1 score and overall F1 score, in MS-COCO, for all methods in the appendix (Appx.~\ref{sec:additional}).
For class-incremental learning, we show the accuracy of methods after training on the last task and evaluating all seen classes. For each metric, we highlight in \textbf{bold} the best approach and \underline{underline} the second best.

\begin{table}
    \centering
    \small
    \caption{Comparison with the state-of-the-art on MS-COCO B0-C10 and B40-C10 configurations. We highlight \inlineColorbox{\colormethod}{\name}, \inlineColorbox{\colorcil}{prompt-based methods}, and \inlineColorbox{\colormlcil}{MLCIL approaches}.}
    \begin{tabular}{l c c c c c}
    \toprule
    \multirow{2.5}{*}{\textbf{Method}} & \multirow{2.5}{*}{\textbf{Buffer size}} & \multicolumn{2}{c}{\textbf{MS-COCO B0-C10}} & \multicolumn{2}{c}{\textbf{MS-COCO B40-C10}} \\
    \cmidrule(lr){3-6}
        & & Avg. mAP $(\uparrow)$ & mAP $(\uparrow)$ & Avg. mAP $(\uparrow)$ & mAP $(\uparrow)$ \\
    \midrule
        UB & - & - & 81.8 & - & 81.8 \\
    \cmidrule{1-6}
        FT~\citep{ridnik2021asymmetric} & \multirow{6}{*}{0} & 38.3 & 16.9 & 35.1 & 17.0 \\
        PODNet~\citep{douillard2020podnet} & & 43.7 & 25.6 & 44.3 & 24.7 \\
        oEWC~\citep{schwarz2018progress} & & 46.9 & 24.3 & 44.8 & 27.3 \\
        LWF~\citep{li2017learning} & & 47.9 & 28.9 & 48.6 & 29.9 \\
        \rowmlcil
        KRT~\citep{dong2023knowledge} & & 74.6 & 65.9 & 77.8 & 74.0 \\
        \rowcil
        CODA-P~\citep{smith2023coda} & & 74.0 & 65.4 & 73.9 & 67.5 \\
    \cmidrule{1-6}
        iCaRL~\citep{rebuffi2017icarl} & \multirow{7}{*}{20/class} & 59.7 & 43.8 & 65.6 & 55.7 \\
        BiC~\citep{wu2019large} & & 65.0 & 51.1 & 65.5 & 55.9 \\
        ER~\citep{DBLP:conf/iclr/RiemerCALRTT19} & & 60.3 & 47.2 & 68.9 & 61.6 \\
        TPCIL~\citep{tao2020topology} & & 69.4 & 60.6 & 72.4 & 66.5 \\
        PODNet~\citep{douillard2020podnet} & & 70.0 & 58.8 & 71.0 & 64.2 \\
        DER++~\citep{buzzega2020dark} & & 72.7 & 63.1 & 73.6 & 66.3 \\
        \rowmlcil
        KRT-R~\citep{dong2023knowledge} & & \underline{76.5} & \underline{70.2} & \underline{78.3} & \underline{75.2} \\
    \cmidrule{1-6}
        \rowmlcil
        PRS~\citep{kim2020imbalanced} & & 48.8 & 27.9 & 50.8 & 33.2 \\
        \rowmlcil
        OCDM~\citep{liang2022optimizing} & 1000 & 49.5 & 28.5 & 51.3 & 34.0 \\
        \rowmlcil
        KRT-R~\citep{dong2023knowledge} & & 75.7 & 69.3 & \underline{78.3} & 75.1 \\
    \cmidrule{1-6}
        \rowmethod
        \textbf{\name (Ours)} & 0 & \textbf{79.1} & \textbf{74.5} & \textbf{78.8} & \textbf{76.6} \\
    \bottomrule
    \end{tabular}
    \label{tab:coco}
\end{table}

\paragraph{Baselines.}
We focus our comparison with the current state-of-the-art in multi-label class-incremental learning, KRT~\citep{dong2023knowledge}.
Because the literature is relatively unexplored, we also evaluate our approach against online MLCIL methods, PRS~\citep{kim2020imbalanced}, and OCDM~\citep{liang2022optimizing}.
We proceed by comparing \name with rehearsal-free approaches that are not designed for MLCIL, CODA-Prompt~\citep{smith2023coda}, Learning without Forgetting~\citep{li2017learning}, and oEWC~\citep{schwarz2018progress}.
Additionally, we provide a comparison with rehearsal-based continual learning techniques, ER~\citep{DBLP:conf/iclr/RiemerCALRTT19}, BiC~\citep{wu2019large}, DER~\citep{buzzega2020dark}, PODNet~\citep{douillard2020podnet}, TPCIL~\citep{tao2020topology}, and iCaRL~\citep{rebuffi2017icarl} and we evaluate them using a replay buffer of 2/20 samples per class, and 1000 exemplars, based on the benchmark dataset.

\paragraph{Implementation details.}
Following the previous works in prompt-based continual learning, we used ViT-B/16, pre-trained on ImageNet-1k, with patches of dimension 16 $\times$ 16.
To train it, we used Adam optimizer~\citep{DBLP:journals/corr/KingmaB14} with $\beta_1=0.9$, and $\beta_2=0.999$ and a batch size of 128 and 256 images for MS-COCO and VOC2007 respectively.
All experiments were executed using a single A100 NVIDIA GPU.
The learning rate followed a cosine annealing schedule, starting from 0.03 for MS-COCO and 0.05 for VOC2007.
Input images were scaled to a dimension of 224 $\times$ 224, which corresponds to the default ViT-B/16 input dimension.

About the hyperparameters strictly related to our method, we set the number of Patch Selectors and the number of prompts to 20 for MS-COCO, and 10 for VOC2007, as we saw no significant improvements in increasing them over these values (see Fig.~\ref{fig:tome}).
We outline that using a standard ViT-B/16, each Patch Selector adds $D=768$ parameters, thus, the total Patch Selector parameters number is counted as $T\times L_\vs \times D$ (\textit{e.g.} for 8 tasks and 20 Patch Selectors, we add 122'880 parameters).
Following~\citep{wang2022dualprompt, smith2023coda}, we used prefix-tuning, and prompts were injected in the first five layers of ViT. %

\subsection{Comparison with the State-of-the-Art}
\label{sub:comparison}
In Table~\ref{tab:coco} and Table~\ref{tab:voc}, we compare \name with the baselines in respectively the MS-COCO~\citep{lin2014microsoft} and VOC2007~\citep{pascal-voc-2007} datasets (see Appx.~\ref{sec:additional} for additional metrics).

\name achieves a new state of the art in both scenarios we evaluated without storing samples for future replay.
The first scenario we consider is incremental learning over the entire MS-COCO dataset, consisting of 8 tasks.
Here, our method outperforms KRT-R's best mAP by 4.3\% and achieves an Average mAP of 79.1\% (+2.6\% improvement), thus, highly reducing the gap with the joint-learning upper bound. 
However, in the 5-tasks B40-C10 configuration\footnote{To learn the base classes and then the incremental steps, we treated the base classes as a bigger task.}, we surpass the final mAP of the previous state-of-the-art by 1.5\% and the Average one by about 0.5\%.
As the number of tasks reduces, the chance that new images are negative samples for old classes is strongly reduced, thus, also the benefit of splitting the forward pass into multiple pathways.
Moreover, in MS-COCO B40-C10, the first task contains approximately half of the dataset, forcing our method to rely on 20 summarized patches to predict 40 different classes.
Despite being disadvantaged, our method still achieves a new state of the art in this configuration.

In VOC2007 our approach improves CODA-P performances (previous state-of-the-art) mAP and Average mAP by about 4.3\% and 2.3\% respectively, getting closer to the upper bound.
Contrary to what happens in MS-COCO B40-C10, we maintain a larger margin from the previous SOTA in the 6-tasks B10-C2 VOC2007 configuration.
Compared to MS-COCO B40-C10, where the number of tasks drops from eight to five, the number of tasks of VOC2007 B10-C2 grows from 5 to 6.
Thus, the likelihood of an image becoming a negative sample remains almost identical.
Additionally, having a lot of tasks with a low number of classes allows our model to exploit the multiple pathways even better, making incremental tasks easier.

Compared with previous approaches, we finally notice \name robustness across different datasets and configurations, highlighting the benefits of designing a prompt-based approach that relies on all trained parameters, an essential feature in MLCIL.
Furthermore, we remark on the benefits of using a multiple \textit{pathways} architecture in long-tail distributed datasets in Appx.~\ref{sec:coda}.

\begin{table}
    \centering
    \small
    \caption{Comparison with the state-of-the-art on VOC2007 B0-C4 and B10-C2 configurations. We highlight \inlineColorbox{\colormethod}{\name}, \inlineColorbox{\colorcil}{prompt-based methods}, and \inlineColorbox{\colormlcil}{MLCIL approaches}.}
    \begin{tabular}{l c c c c c}
    \toprule
        \multirow{2.5}{*}{\textbf{Method}} & \multirow{2.5}{*}{\textbf{Buffer size}} & \multicolumn{2}{c}{\textbf{VOC2007 B0-C4}} & \multicolumn{2}{c}{\textbf{VOC2007 B10-C2}} \\
        \cmidrule(lr){3-6}
         & & Avg. mAP $(\uparrow)$ & mAP $(\uparrow)$ & Avg. mAP $(\uparrow)$ & mAP $(\uparrow)$ \\
    \midrule
        UB & - & - & 93.6 & - & 93.6 \\
        \rowcil
        CODA-P~\citep{smith2023coda} & 0 & 90.6 & \underline{84.5} & \underline{90.2} & \underline{85.0} \\
    \cmidrule{1-6}
        FT~\citep{ridnik2021asymmetric} & \multirow{8}{*}{2/class} & 82.1 & 62.9 & 70.1 & 43.0 \\
        iCaRL~\citep{rebuffi2017icarl} & & 87.2 & 72.4 & 79.0 & 66.7 \\
        BIC~\citep{wu2019large} & & 86.8 & 72.2 & 81.7 & 69.7 \\
        ER~\citep{DBLP:conf/iclr/RiemerCALRTT19} & & 86.1 & 71.5 & 81.5 & 68.6 \\
        TPCIL~\citep{tao2020topology} & & 87.6 & 77.3 & 80.7 & 70.8 \\
        PODNet~\citep{douillard2020podnet} & & 88.1 & 76.6 & 81.2 & 71.4 \\
        DER++~\citep{buzzega2020dark} & & 87.9 & 76.1 & 82.3 & 70.6 \\
        \rowmlcil
        KRT-R~\citep{dong2023knowledge} & & \underline{90.7} & 83.4 & 87.7 & 80.5 \\
    \cmidrule{1-6}
        \rowmethod
        \textbf{\name (Ours)} & 0 & \textbf{93.5} & \textbf{88.8} & \textbf{93.1} & \textbf{88.3} \\
    \bottomrule  
    \end{tabular}
    \label{tab:voc}
\end{table}

\subsection{Qualitative Results}
\label{sub:qualitative}
\begin{figure}
    \centering
    \includegraphics[width=\linewidth]{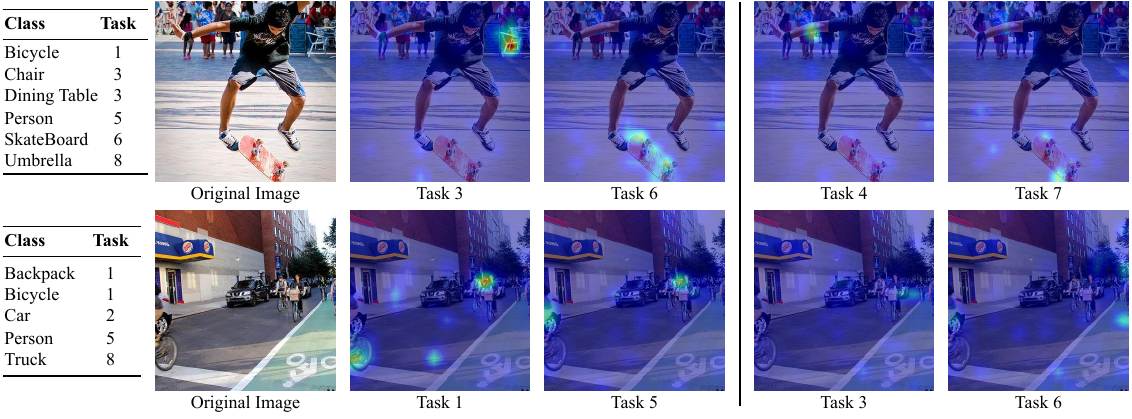}
    \caption{Visualization of the attention paid by Patch Selectors of different tasks. On the left, are ground truth annotations for images. On the right, are depicted original images, two heatmaps for tasks related to the classes of the image, and two other heatmaps for tasks that are not related to the classes of the image.}
    \label{fig:heatmap}
\end{figure}
As explained in Sec.~\ref{sec:method}, for each task \name learns a set of Patch Selectors that focus on specific parts of the image to produce a summarization tailored to the specific task.
In Fig.~\ref{fig:heatmap}, we present a visualization of the attention paid by Patch Selectors of different incremental steps.
In particular, we show side by side two heatmaps produced by tasks containing classes that are in the image (left) and two for tasks whose classes are not (right).
In the first row of Fig.~\ref{fig:heatmap}, we can appreciate how Patch Selectors of task 3 focus on the region where the chair and the dining table happen to be. 
Analogously, task 6 Patch Selectors are summarizing the image by giving the most attention to the skateboard.
Despite focusing on important aspects of the scene, Patch Selectors seem to be also activated by shadows or elements that could be in the scene.
An example of such behavior can be seen in the heatmaps of tasks 4 and 7 of the first row (whose classes are not in the image); where they show peaked activation in regions that are not directly relevant in predicting correct classes.
We hypothesize that task 4 and task 7 strong activations may be generated by Patch Selectors focusing on finding respectively a handbag and a snowboard\footnote{Classes that belong to respectively task 4 and 7.}.
In the second row, instead, Patch Selectors seem to activate also in the presence of shadows or certain patterns in the terrain.
Thus, we argue that they do not only activate in the presence of a specific object but also activate to retrieve context from the scene (\textit{e.g.} by looking at regions that are usually populated by certain objects or patterns).
We show additional failure cases in Appx.~\ref{sec:failure}.

\subsection{Ablations}
\label{sub:ablations}
\paragraph{Components.}
In Table~\ref{tab:ablations} we ablate the components of our approach by averaging the final mAP on MS-COCO B0-C10 across 3 runs with different class orderings and seeds.
We set as a reference \name trained with 20 Patch Selectors per task, Drop \& Replace, pre-head normalization, and multiple \textit{pathways}, as described in Sec.~\ref{sub:compression}. 
We begin our analyses by removing the pre-head normalization, which causes \name performance to drop by about 2\% from the reference.
Then, we test our method without the multiple \textit{pathways} structure we proposed.
\begin{wrapfigure}[14]{r}{.4\textwidth}
    \vspace{-17pt}
    \begin{center}
        \includegraphics[width=\linewidth]{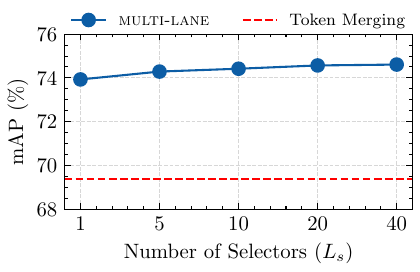}
    \end{center}
    \caption{Patch Selectors vs Token Merging in MS-COCO B0-C10.}
    \label{fig:tome}
\end{wrapfigure}
We instead employed a query-key selection mechanism, typical of prompt-based approaches~\citep{wang2022dualprompt}.
We used frozen forward features as a query vector and selected the most probable task-specific parameters (prompts and Patch Selectors) to compute the final prediction for the image.
By using only one \textit{pathway} our method suffers from the same architectural issues of prompt-based approaches (see Sec.~\ref{sub:clpt}), causing a performance drop.
Finally, we ablate the Drop\&Replace operation.
Instead of summarizing patch tokens at each ViT block, we summarize them only at the first layer and propagate them, causing the mAP to drop to $\sim$20\% (see Fig.~\ref{fig:hetamap_all} in the Appendix).

\paragraph{Patch Summarization.}
In Fig.~\ref{fig:tome}, we investigate the importance of Patch Selectors in our architecture.
In detail, we show the performance of our model when varying the number of Patch Selectors $L_s$.
Additionally, we show \name performance when Token Merging (ToME) \citep{bolya2022tome} is used to summarize patch tokens (details in Appx.~\ref{sec:tome}). 
ToMe proves to be suboptimal as it does not isolate the foreground from the background for a specific task.
On the other hand, Patch Selectors are more effective as they are tailored for the task they have been trained on.
Moreover, we notice how the performances are not much affected by the number of Patch Selectors, demonstrating that enough information can be encoded even into a single patch.

\begin{table}
    \parbox[b]{.40\linewidth}{
        \centering
        \small
        \caption{Component ablations of our approach. In the last row, we report \inlineColorbox{\colormethod}{\name} with all 3 components enabled.}
        \begin{tabular}{c c c c}
        \toprule
            \multicolumn{3}{c}{\textbf{Component}} & \multirow{2.5}{*}{\textbf{mAP} $(\uparrow)$} \\
            \cmidrule(lr){1-3}
             Norm & Parallel \textit{Pathways} &  D\&R & \\
        \midrule
            & \scalecheck & \scalecheck & 72.66\scriptsize$\pm0.18$ \\
            \scalecheck & & \scalecheck & 67.96\scriptsize$\pm0.30$ \\
            \scalecheck & \scalecheck & & 20.26\scriptsize$\pm0.36$ \\
            \rowmethod
            \scalecheck & \scalecheck & \scalecheck & 74.57\scriptsize$\pm0.06$ \\
        \bottomrule
        \end{tabular}
        \label{tab:ablations}
    }
    \hfill
    \parbox[b]{.55\linewidth}{
        \centering
        \small
        \caption{Comparison in terms of Giga Multiply-Accumulate operations, frames per second (FPS), and trainable parameters on a 10-task benchmark. Patch Selectors drastically reduce the computational overhead of a na\"ive multiple \textit{pathways} architecture.}
        \begin{tabular}{l c c c}
        \toprule
            \textbf{Approach} & GMACs $(\downarrow)$ & FPS ($\uparrow$) & \makecell{Trainable\\Params (M)} \\
        \midrule
            ViT-B/16 & 16.9 & 918 & 85.8 \\
            Na\"ive multiple \textit{pathways} & 168.7 & 92 & 858.0 \\
            \rowcil
            CODA-P & 33.7 & 436 & 3.7 \\
        \cmidrule{1-4}
            \rowmethod
            \name{} - 1 Selector & 18.6 & 790 & 1.2 \\
            \rowmethod
            \name{} - 20 Selectors & 34.7 & 393 & 1.3 \\
        \bottomrule
        \end{tabular}
        \label{tab:macs}
    }
\end{table}

\subsection{Computational requirements}
In Sec.~\ref{sub:compression}, we provide a theoretical perspective on the computational demands of the multiple \textit{pathways} architecture and how Patch Selectors can solve it.
To further demonstrate the validity of our analyses, we empirically demonstrate that Patch Selectors do strongly reduce the computational requirements. 
In Table~\ref{tab:macs}, we report the number of multiply-accumulate operations, the frames per second, and the number of trained parameters required by our method on 10-task incremental sessions and we compare them with CODA-Prompt, a Na\"ive implementation of multiple \textit{pathways}, and the unaltered ViT-B/16 forward.
On a 10-task benchmark \name requires GMACs operations and FPS similar to a ViT-B/16 forward while being capable of achieving a new state-of-the-art in MS-COCO B0-C10. Note that MS-COCO B0-C10 is divided into 8 tasks, thus, it requires slightly fewer GMACs for this setting. %
By raising the number of Patch Selectors to 20, our approach reaches computational requirements close to CODA-P.
Furthermore, we want to stress that, in their lighter configuration, Patch Selectors require almost 10$\times$ fewer GMAC operations and are almost 10$\times$ faster compared to Na\"ive multiple \textit{pathways}.
This proves that, compared to the previous state-of-the-art, our methodology is not only superior in mAP terms but also as "light" as a plain ViT-B/16.

\subsection{Additional results in CIL}
\label{sub:additional}
To show the versatility of our method, we further evaluate our approach in the class-incremental learning setting (see Table~\ref{tab:cil}).
We compare \name against the state-of-the-art prompt-based class-incremental learning approaches in the CIFAR-100 and ImageNet-R benchmarks. Although our method is designed for MLCIL, \name is also robust in class-incremental learning, as it consistently outperforms seminal prompt-based approaches~\citep{wang2022learning, wang2022dualprompt}.
However, \name is not designed for CIL and it is not capable of reaching the same performance as CODA-Prompt, which is the SOTA in prompt-based CIL.
Finally, we show that methods designed for MLCIL may not be directly applied to CIL.
Indeed, KRT~\citep{dong2023knowledge} fails in achieving meaningful performances in this setting.

\begin{table*}
    \centering
    \small
    \caption{Comparison with the state-of-the-art in popular class-incremental learning benchmarks. We highlight \inlineColorbox{\colormethod}{\name}, \inlineColorbox{\colorcil}{prompt-based methods}, and \inlineColorbox{\colormlcil}{MLCIL approaches}.}
    \begin{tabular}{l c c c c}
    \toprule
        \multirow{2.5}{*}{\textbf{Method}} & \multirow{2.5}{*}{\textbf{Setting}} & \multirow{2.5}{*}{\textbf{Buffer size}} & \multicolumn{2}{c}{\textbf{Accuracy} $(\uparrow)$}\\
        \cmidrule(lr){4-5}
         & & & CIFAR-100 & ImageNet-R \\
    \midrule
        UB & - & - & $90.85$\scriptsize$\pm0.12$ & $79.13$\scriptsize$\pm0.18$ \\
    \cmidrule{1-5}
        \rowmlcil
        KRT-R~\citep{dong2023knowledge} & MLCIL & 5000 & 65.10\scriptsize$\pm2.60$ & 63.67\scriptsize$\pm1.27$ \\
        \rowcil
        L2P~\citep{wang2022learning} & & & $82.50$\scriptsize$\pm1.10$ & $69.29$\scriptsize$\pm0.73$ \\
        \rowcil
        DualPrompt~\citep{wang2022dualprompt} & CIL & 0 & $83.05$\scriptsize$\pm1.16$ & $71.31$\scriptsize$\pm0.62$ \\
        \rowcil
        CODA-P~\citep{smith2023coda} & & & \textbf{86.25}\scriptsize$\pm0.74$ & \textbf{75.45}\scriptsize$\pm0.56$ \\
    \cmidrule{1-5}
        \rowmethod
        \name (Ours) & MLCIL & 0 & \underline{84.84}\scriptsize$\pm0.34$ & \underline{73.69}\scriptsize$\pm0.40$ \\
    \bottomrule
    \end{tabular}
    \label{tab:cil}
\end{table*}

\section{Conclusions}
\label{sec:conclusions}
In this work, we present \name, a novel prompt-based approach that deals with multi-label class-incremental.
Contrary to previous prompt-based works that output a single representation for each image, we propose to compute multiple representations in parallel.
By outputting multiple feature vectors, our method can put more focus on each salient point of the image.
To avoid the computational complexity from growing too much, we introduce a novel component we call Patch Selectors that heavily summarizes the internal representation of the ViT.
Patch Selectors allow our method to generate a feature vector for each encountered task dataset, limiting forgetting.
Experiments in MS-COCO and VOC2007 datasets show that \name achieves a new state-of-the-art in multi-label class-incremental learning.

\subsubsection*{Acknowledgments}
We acknowledge the CINECA award under the ISCRA initiative for the availability of high-performance computing resources and support. E.R. and M.M. are supported by the MUR PNRR project FAIR - Future AI Research (PE00000013), funded by NextGeneration EU. E.R. is also supported by the EU projects AI4TRUST (No.101070190) and ELIAS (No.01120237) and the PRIN project LEGO-AI (Prot.2020TA3K9N).
T.D.M. is funded by NextGeneration EU.
This work has been supported by the French National Research Agency (ANR) with the ANR-20-CE23-0027.

\bibliography{collas2024_conference}
\bibliographystyle{collas2024_conference}
\appendix
\newpage
{\centering\Large\bf%
 Appendix%
}

\section{Task Transition}
\label{sec:transition}

\begin{figure}[h]
    \centering
    \includegraphics[width=.66\linewidth]{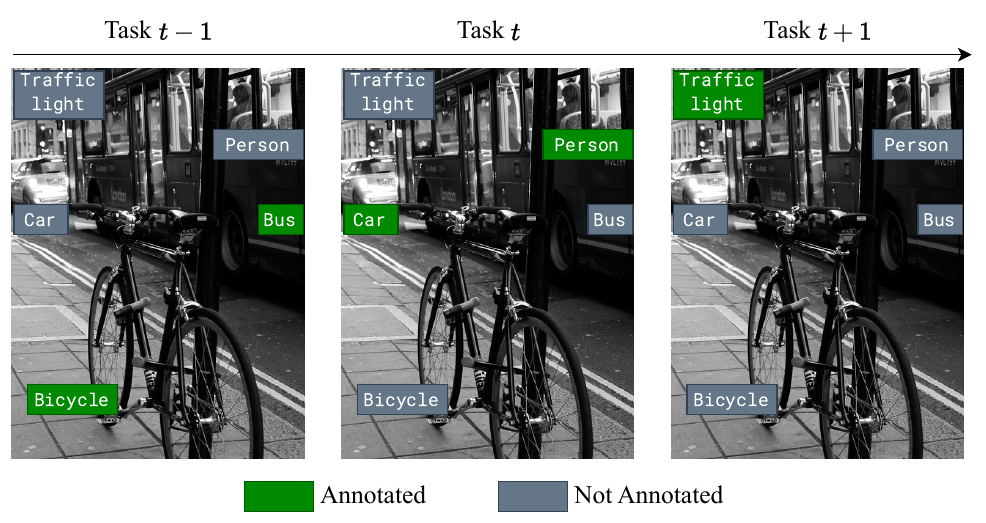}
    \caption{\textbf{Annotations across task transitions}: Past task objects can appear in future tasks in unlabeled form.}
    \label{fig:transition}
\end{figure}

In this paragraph we provide an overview of the task transition in MLCIL, highlighting how negative samples for past tasks are formed.
In class-incremental learning, object categories appear only in one task during the incremental learning (\textit{e.g.} if the class bicycle is observed in task N, no image with bicycle annotated can appear in task M, with $M\neq N$).
However, real-world images rarely depict only one class: thus, multi-label class-incremental learning allows images to belong to multiple categories. Following standard definitions for this task~\citep{dong2023knowledge}, a class is \textit{annotated} only in one incremental step while it can still appear but \textit{unlabeled} in other steps. %
In Fig.~\ref{fig:transition}, we show a toy model of the task transition just described. 
In task $t-1$, only \textit{bicycle} and \textit{bus} are annotated, leaving the other three classes as background.
Moving to task $t$, previously annotated classes are now considered background, and only \textit{car} and \textit{person} are annotated.
As a result, task $t$ image becomes a negative sample for task $t-1$ classes.
Similarly, for task $t+1$, labels are provided only for the traffic light class, making it a negative sample for previous incremental steps. Note that, for simplicity, in Fig.~\ref{fig:transition} we considered the same image with different annotations. In practice, no assumption is made on the overlap between sets of images across different tasks.

\section{Prompt selection issues in MLCIL}
\label{sec:misinterpret}
\definecolor{centroid}{RGB}{255, 179, 102}
\definecolor{taskv}{RGB}{102, 178, 255}
\definecolor{taskt}{RGB}{255, 102, 102}
\begin{wrapfigure}[16]{r}{.5\textwidth}
    \vspace{-12pt}
    \centering
    \includegraphics[width=\linewidth]{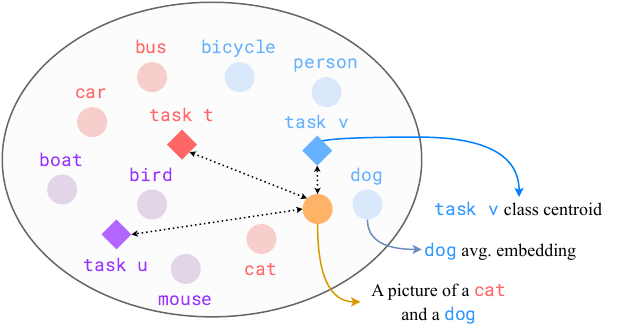}
    \caption{\textbf{Prompt selection}: The \inlineColorbox{centroid}{query} vector that embeds an image of a \textit{cat} and a \textit{dog} is close to \inlineColorbox{taskv}{task $v$} but distant to \inlineColorbox{taskt}{task $t$} in the embedding space. This distance could result in a missed detection of the class \textit{cat}.}
    \label{fig:misinterpret}
\end{wrapfigure}
In Sec.~\ref{sec:introduction}, we argue that prompt-tuning-based methods are more prone to select incorrect prompts in MLCIL than in CIL, as distinct foreground objects require different prompts to be selected.
In this section, we elaborate on this with the help of a toy visualization shown in Fig.~\ref{fig:misinterpret}.

At inference, given an image containing a cat and a dog, the visual encoder will extract embeddings close to the ones of \textit{cat} and \textit{dog} concepts.
We refer to such embedding as the query. Task prompts are represented by task centroids (or keys), which lie in the same latent space as queries.
By computing the cosine similarity between the query and all the keys, prompt-based methods either do a hard~\citep{wang2022dualprompt} or a soft prompt selection~\citep{wang2022learning,smith2023coda}.

A hard prompt selection, as done in~\citep{wang2022dualprompt}, will result in a prompt corresponding to only a subset of the image content. In the example shown in Fig.~\ref{fig:misinterpret}, \textit{cat} is part of task $t$, while \textit{dog} of task $v$. However, only task $v$ key would be selected, being the nearest key from the query, leading to a prompt selection not specialized for \textit{cats}.
While a soft selection mechanism remedies the omission of relevant prompts by considering all task prompts, their importance is weighted proportionally to the query-task centroid distance~\citep{smith2023coda}.
However, the task centroids may not faithfully represent the semantic relevance of a task (e.g. task $t$ contains \textit{cat} but its centroid is influenced by other classes such as \textit{bus} and \textit{car}). %
Moreover, unrelated tasks (e.g. task $u$ in Fig.~\ref{fig:misinterpret}) may have a similar distance with the query as task $t$, and thus, it could interfere with the final classification. Contrarily, our approach sidesteps the issues of representing tasks by task centroids and their selection altogether by extracting task-specific representations in parallel.

\begin{figure}
    \centering
    \includegraphics[width=\linewidth]{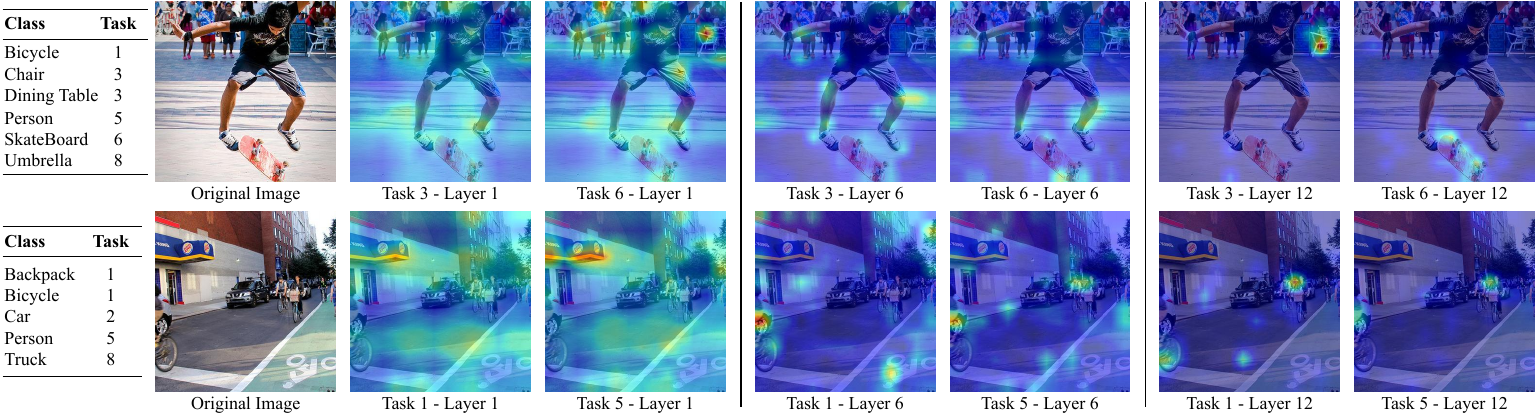}
    \caption{\textbf{Visualizing the summarization at different layers}: The summarization is refined at each step as representations of the frozen pre-trained model become more high-level, better capturing the image semantics. In layer 1, heatmaps tend to be uniform and quite homogeneous among tasks, suggesting that information is not rich enough to provide meaningful summarizations. In layer 6, the specificity of the summarized patches increases, starting to focus on more discriminative elements of the image (e.g. \textit{skateboard} and \textit{human} in the first row, \textit{road} and \textit{person} in the second). Finally, in layer 12, Patch Selectors focus on specific image regions, retaining only the semantic information (e.g. \textit{skateboard} in the first row, \textit{bicycle} and \textit{person} in the second).}
    \label{fig:hetamap_all}
\end{figure}

\section{Additional results in MS-COCO}
\label{sec:additional}
In Table~\ref{tab:coco_full}, we show results with the same experimental procedure of Table~\ref{tab:coco} with two additional metrics: the class-wise F1 score and the overall F1 score.
We did not search for the optimal threshold for both metrics and used 0.8 as a decision boundary like in \citet{dong2023knowledge}.
Indeed, \name does not achieve state-of-the-art results, in such metrics, in three out of four cases.
However, we consider mAP and Avg. mAP more descriptive metrics and they are hyperparameter-free, thus, we focused our analyses on those and we neglected the other two.
Here we report them for completeness.

\begin{table}
    \centering
    \scriptsize
    \caption{Comparison with the state-of-the-art on MS-COCO B0-C10 and B40-C10 configurations. We highlight \inlineColorbox{\colormethod}{\name}, \inlineColorbox{\colorcil}{prompt-based methods}, and \inlineColorbox{\colormlcil}{MLCIL approaches}.}
    \begin{tabular}{l c c c c c c c c c}
    \toprule
        \multirow{2.5}{*}{\textbf{Method}} & \multirow{2.5}{*}{\textbf{Buffer size}} & \multicolumn{4}{c}{\textbf{MS-COCO B0-C10}} & \multicolumn{4}{c}{\textbf{MS-COCO B40-C10}} \\
    \cmidrule(lr){3-10}
          & & Avg. mAP $\uparrow$ & CF1 $\uparrow$ & OF1 $\uparrow$ & mAP $\uparrow$ & Avg. mAP $\uparrow$ & CF1 $\uparrow$ & OF1 $\uparrow$ & mAP $\uparrow$ \\
    \midrule
        UB & - & - & 76.4 & 79.4 & 81.8 & - & 76.4 & 79.4 & 81.8 \\
    \cmidrule{1-10}
        FT~\citep{ridnik2021asymmetric} & \multirow{6}{*}{0} & 38.3 &  6.1 & 13.4 & 16.9 & 35.1 & 6.0 & 13.6 & 17.0 \\
        PODNet~\citep{douillard2020podnet} & & 43.7 &  7.2 & 14.1 & 25.6 & 44.3 & 6.8 & 13.9 & 24.7 \\
        oEWC~\citep{schwarz2018progress} & & 46.9 &  6.7 & 13.4 & 24.3 & 44.8 & 11.1 & 16.5 & 27.3 \\
        LWF~\citep{li2017learning} & & 47.9 &  9.0 & 15.1 & 28.9 & 48.6 & 9.5 & 15.8 & 29.9 \\
        \rowmlcil
        KRT~\citep{dong2023knowledge} & & 74.6 &  55.6 & 56.5 & 65.9 & 77.8 & 64.4 & 63.4 & 74.0 \\
        \rowcil
        CODA-P~\citep{smith2023coda} & & 74.0 & 48.1 & 47.7 & 65.4 & 73.9 & 56.1 & 57.4 & 67.5 \\
    \cmidrule{1-10}
        TPCIL~\citep{tao2020topology} & \multirow{4}{*}{5/class} & 63.8 & 20.1 & 21.6 & 50.8 & 63.1 & 25.3 & 25.1 & 53.1 \\
        PODNet~\citep{douillard2020podnet} & & 65.7 & 13.6 & 17.3 & 53.4 & 65.4 & 24.2 & 23.4 & 57.8 \\
        DER++~\citep{buzzega2020dark} & & 68.1 & 33.3 & 36.7 & 54.6 & 69.6 & 41.9 & 43.7 & 59.0 \\
        \rowmlcil
        KRT-R~\citep{dong2023knowledge} & & 75.8 & 60.0 & 61.0 & 68.3 & 78.0 & 66.0 & 65.9 & 74.3 \\
    \cmidrule{1-10}
        iCaRL~\citep{rebuffi2017icarl} & \multirow{7}{*}{20/class} & 59.7 & 19.3 & 22.8 & 43.8 & 65.6 & 22.1 & 25.5 & 55.7 \\
        BiC~\citep{wu2019large} & & 65.0 & 31.0 & 38.1 & 51.1 & 65.5 & 38.1 & 40.7 & 55.9 \\
        ER~\citep{DBLP:conf/iclr/RiemerCALRTT19} & & 60.3 & 40.6 & 43.6 & 47.2 & 68.9 & 58.6 & 61.1 & 61.6 \\
        TPCIL~\citep{tao2020topology} & & 69.4 & 51.7 & 52.8 & 60.6 & 72.4 & 60.4 & 62.6 & 66.5 \\
        PODNet~\citep{douillard2020podnet} & & 70.0 & 45.2 & 48.7 & 58.8 & 71.0 & 46.6 & 42.1 & 64.2 \\
        DER++~\citep{buzzega2020dark} & & 72.7 & 45.2 & 48.7 & 63.1 & 73.6 & 51.5 & 53.5 & 66.3 \\
        \rowmlcil
        KRT-R~\citep{dong2023knowledge} & & \underline{76.5} & \underline{63.9} & \textbf{64.7} & \underline{70.2} & \underline{78.3} & \textbf{67.9} & \textbf{68.9} & \underline{75.2} \\
    \cmidrule{1-10}
        \rowmlcil
        PRS~\citep{kim2020imbalanced} & & 48.8 & 8.5 & 14.7 & 27.9 & 50.8 & 9.3 & 15.1 & 33.2 \\
        \rowmlcil
        OCDM~\citep{liang2022optimizing} & 1000 & 49.5 & 8.6 & 14.9 & 28.5 & 51.3 & 9.5 & 15.5 & 34.0 \\
        \rowmlcil
        KRT-R~\citep{dong2023knowledge} & & 75.7 & 61.6 & \underline{63.6} & 69.3 & \underline{78.3} & \underline{67.5} & \underline{68.5} & 75.1 \\
    \cmidrule{1-10}
        \rowmethod
        \textbf{\name (Ours)} & 0 &\textbf{79.1} & \textbf{65.1} & 62.8 & \textbf{74.5} & \textbf{78.8} & 66.0 & 66.6 & \textbf{76.6} \\
    \bottomrule
    \end{tabular}
    \label{tab:coco_full}
\end{table}

\section{\name pseudocode}
\label{sec:pseudo}
In Algo.~\ref{alg:summarization}, we report the pseudocode of the multi-head attention of \name in a PyTorch-like style.
First, we compute the patch tokens summarization and concatenate summarized patches to task-specific class tokens.
We compute the query, key, and value projections, and, if provided, we concatenate prompts to key and value matrices.
The three projections are used to compute the task MSA, and summarized patches are dropped.
Finally, we use the original patch tokens to compute the MSA for the frozen forward. 

\definecolor{comment}{RGB}{40, 106, 191}
\definecolor{reserved}{RGB}{207, 27, 63}
\definecolor{string}{RGB}{30, 158, 2}
\definecolor{fun}{RGB}{118, 38, 209}
\newcommand{\comm}[1]{\textcolor{comment}{\# #1}}
\newcommand{\reserved}[1]{\textcolor{reserved}{#1}}
\newcommand{\str}[1]{\textcolor{string}{#1}}
\newcommand{\fun}[1]{\textcolor{fun}{#1}}
\begin{algorithm}
    \caption{\name Multi-head Attention}
    \label{alg:summarization}
    \begin{algorithmic}
        \tt
        \small
        \State \reserved{def} \fun{summarize\_patches}(patches, patch\_selectors):
        \State \qquad \comm{Input:}
        \State \qquad \comm{ - patches: [batch\_size, num\_tokens, token\_dim]}
        \State \qquad \comm{ - patch\_selectors: [num\_tasks, num\_selectors, token\_dim]}
        \State \qquad \comm{Output:}
        \State \qquad \comm{ - summarized\_patches: [num\_tasks, batch\_size, num\_selectors, token\_dim]}
        \State \qquad attention\_scores \reserved{=} \fun{einsum}(\str{"tcd, bnd -> tbcn"}, patch\_selectors, \fun{detach}(patches))
        \State \qquad norm\_scores \reserved{=} \fun{softmax}(attention\_scores \reserved{*} (token\_dim\reserved{**}-0.5), dim\reserved{=}-1)
        \State \qquad summarized\_patches \reserved{=} \fun{einsum}(\str{"tbcn, bdn -> tbcd"}, norm\_scores, \fun{detach}(patches))
        \State \qquad \reserved{return} summarized\_patches
        \State 
        \State \reserved{def} \fun{multi\_head\_attention}(patches, task\_cls\_tokens, patch\_selectors, prompts:Optional):
        \State \qquad \comm{Input:}
        \State \qquad \comm{ - patches: [batch\_size, num\_tokens, token\_dim]}
        \State \qquad \comm{ - task\_cls\_tokens: [num\_tasks, batch\_size, 1, token\_dim]}
        \State \qquad \comm{ - patch\_selectors: [num\_tasks, num\_selectors, token\_dim]}
        \State \qquad \comm{ - prompts: [2, num\_tasks, batch\_size, num\_heads, length, head\_dim]}
        \State \qquad \comm{Output:}
        \State \qquad \comm{ - patches: [batch\_size, num\_tokens, token\_dim]}
        \State \qquad \comm{ - task\_cls\_tokens: [num\_tasks, batch\_size, 1, token\_dim]}
        \State \qquad summarized\_patches \reserved{=} \fun{summarize\_patches}(patches, patch\_selectors)
        \State \qquad task\_tokens \reserved{=} \fun{cat}((task\_cls\_tokens, summarized\_patches), dim\reserved{=}2)
        \State
        \State \qquad \comm{task\_qkv: [3, num\_tasks, batch\_size, num\_heads, num\_selectors+1, head\_dim]}
        \State \qquad task\_qkv \reserved{=} \fun{qkv\_proj}(task\_tokens)
        \State \qquad task\_q, task\_k, task\_v \reserved{=} \fun{unbind}(task\_qkv, dim\reserved{=}0)
        \State \qquad \reserved{if} prompts:
        \State \qquad \qquad task\_k \reserved{=} \fun{cat}((prompts[0], task\_k), dim\reserved{=}3)
        \State \qquad \qquad task\_v \reserved{=} \fun{cat}((prompts[1], task\_v), dim\reserved{=}3)
        \State \qquad task\_tokens \reserved{=} \fun{MHA}(task\_q, task\_k, task\_v)
        \State \qquad
        \State \qquad \comm{Drop\&Replace: Drop summarized\_patches}
        \State \qquad task\_cls\_tokens \reserved{=} task\_tokens[:, :, 0:1]
        \State \qquad
        \State \qquad \comm{Frozen MHA}
        \State \qquad \reserved{with} \fun{no\_grad}():
        \State \qquad \qquad qkv \reserved{=} \fun{qkv\_proj}(patches)
        \State \qquad \qquad q, k, v \reserved{=} \fun{unbind}(qkv, dim\reserved{=}0)
        \State \qquad \qquad patches \reserved{=} \fun{MHA}(q, k, v)
        \State \qquad \reserved{return} patches, task\_cls\_tokens
    \end{algorithmic}
\end{algorithm}

\section{\name ensures a more equal representation capacity}
\label{sec:coda}
\begin{figure}
    \centering
    \begin{subfigure}[b]{\textwidth}
        \centering
        \includegraphics[width=.8\linewidth]{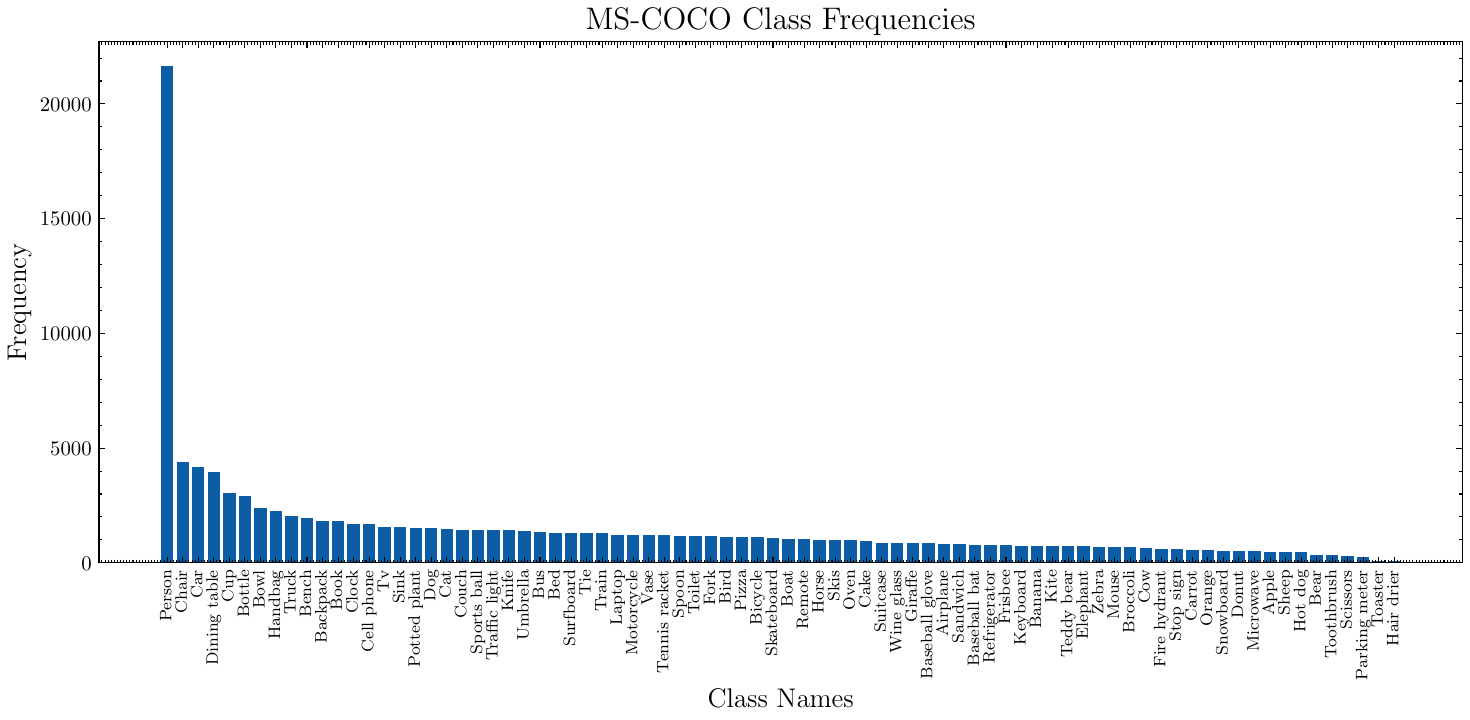}
        \caption{Long tail distribution of MS-COCO.}
        \label{fig:class-histogram}
    \end{subfigure}%
    \vspace{10pt}
    \begin{subfigure}[b]{\textwidth}
        \centering
        \includegraphics[width=.8\linewidth]{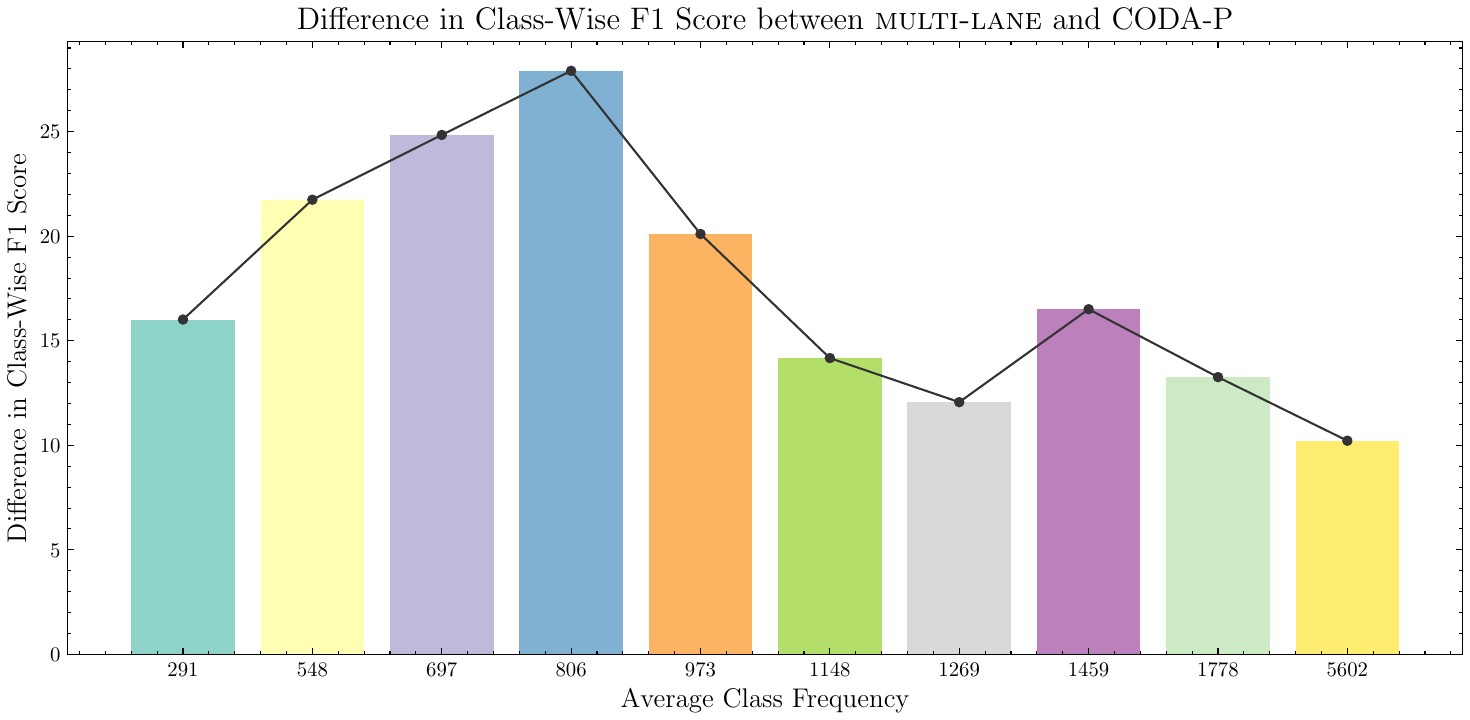}
        \caption{Difference in Class-Wise F1 Score between \name and CODA-P. Classes are grouped based on their frequencies.}
        \label{fig:histogram}
    \end{subfigure}
    \caption{\name is more effective than CODA-P in long-tailed distributed datasets.}
\end{figure}

As we introduce in Sec.~\ref{sec:introduction} and Sec.~\ref{sec:method}, by allocating task-specific forwards, \name provides an equal representation capacity for classes along all tasks.
By assigning different weights to distinct tasks and by exploiting all task-specific parameters in parallel, \name can put more focus on less represented classes.
To demonstrate that multiple \textit{pathways} architecture does mitigate the long-tailed distribution issues of multi-label datasets (see Fig.~\ref{fig:class-histogram}), we computed the difference in class-wise F1 score between our method and CODA-Prompt.
In detail, we computed the F1 score independently for each class, then we averaged the resulting F1 scores and computed the difference between the metric we obtained with \name and with CODA-P.
We grouped classes into 10 bins depending on their frequencies and in Fig.~\ref{fig:histogram} we show the differences between the two methods.
Our method consistently shows better F1 scores, moreover, the difference with CODA-P tends to grow as the number of samples per class lowers.
This clearly shows that \name is more robust to long-tailed distributed datasets.

\section{Token Merging for Summarization}
\label{sec:tome}
In Fig.~\ref{fig:tome}, we present the performance of \name when replacing Patch Selectors with Token Merging (ToMe).
Token Merging \citep{bolya2022tome} works through a bipartite soft matching, where tokens are first split into two sets and then matched based on a similarity measure.
The $r$ most similar pairs are merged and concatenated to the remaining ones.
This means after applying Token Merging, the length of the resulting patches is halved at most.
We aim to reduce the number of summarized patches even further, thus, we applied Token Merging multiple times until the length of the summarized patches was short enough.
In particular, we started from the original patch tokens $\vg\in\mathbb{R}^{(L+1)\times D}$ and applied Token Merging to produce $\bar{\vg}'\in\mathbb{R}^{L_s'\times D}$, then we applied again ToMe to $\bar{\vg}'$ to obtain $\bar{\vg}''\in\mathbb{R}^{L_s''\times D}$ and so on until the final length $L_s$ is less than a specified threshold.
We set the upper bound for $L_s$ to 30. 
Our objective is to strongly reduce the number of patches while retaining as much information as possible.
As for Patch Selectors-generated summarized patches, ToMe ones are dropped after the MSA and calculated in each ViT Block.

\section{Qualitative Results - Failure cases}
\label{sec:failure}
\begin{figure}
    \centering
    \includegraphics[width=\linewidth]{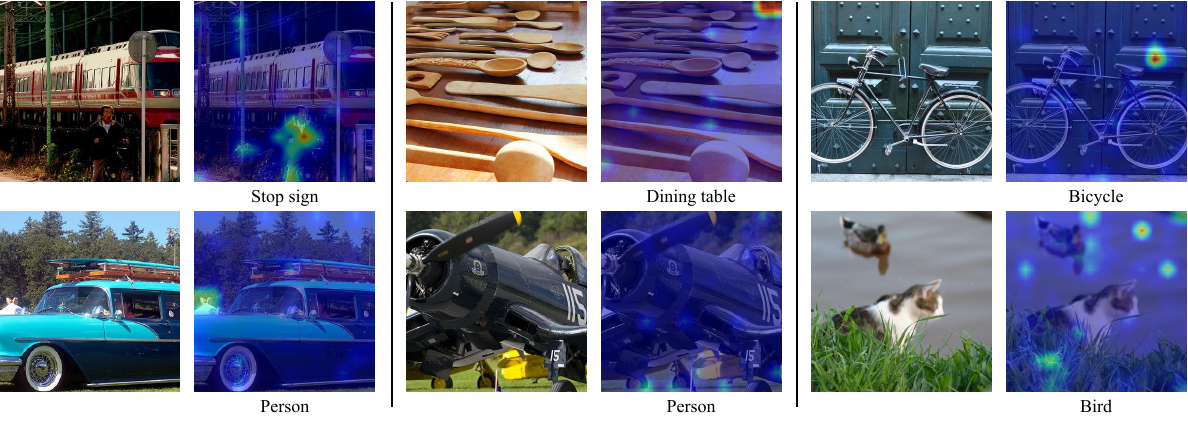}
    \caption{\textbf{Qualitative failure cases}. In these pictures, \name either presents peaked activations in wrong regions of the image, or the activations are very low.}
    \label{fig:failure}
\end{figure}

In Section~\ref{sec:experiments}, we present some qualitative representations of the attention paid by our Patch Selectors when summarizing patch tokens.
In this Section, we report some evident failure cases we encountered when analyzing the heatmaps produced by \name.
First, we noticed that Patch Selectors, especially in the last layers, tend to have peaked activations on specific tokens rather than a more uniform one.
However, when the object takes up most of the image space (\textit{e.g.} airplane picture in Fig.~\ref{fig:failure}), the activation is rather uniform.
When reporting failure cases, we are interested in small objects that do not receive high attention values or the attention is focused on regions not concerning the class \name should look for. 
In Fig.~\ref{fig:failure}, we show six failure pictures with one corresponding task heatmap.
We also show the class for which the Patch Selectors, of the corresponding task, should look.
Finally, we remark that a failure in qualitative results does not imply a failure in classification.

\end{document}